\documentclass[twoside,11pt]{article}

\usepackage{jmlr2eLR}
\usepackage{svg}

\ShortHeadings{Deepchecks: A Library for Testing and Validating Machine Learning Models and Data}{Shir Chorev,  Philip Tannor \emph{et al.}}
\firstpageno{1}

\begin{document}

\title{Deepchecks: A Library for Testing and Validating Machine Learning Models and Data}

\author{\name Shir Chorev \email shir@deepchecks.com\\ \name Philip Tannor \email philip@deepchecks.com \\
\name Dan Ben Israel \email danb@deepchecks.com\\
\name Noam Bressler \email noam@deepchecks.com \\
\name Itay Gabbay 
\email itay@deepchecks.com 
\\
\name Nir Hutnik \email nir@deepchecks.com\\
\name Jonatan Liberman \email jonatan@deepchecks.com\\
\name Matan Perlmutter
\email matan@deepchecks.com 
\\
\name Yurii Romanyshyn 
\email yurii@deepchecks.com
\\
\addr 
Deepchecks Ltd.\\
Derech Menachem Begin 14\\
Ramat Gan, 5270002\\
Israel\\
\AND
\name Lior Rokach,
\email liorrk@bgu.ac.il
\\
\addr Deepchecks Ltd.  and\\
\addr Department of Software and Info. Sys. Eng.,\\ Ben-Gurion University of the Negev.
}

\editor{}

\maketitle

\begin{abstract}%   <- trailing '%' for backward compatibility of .sty file
This paper presents Deepchecks, a Python library for comprehensively validating machine learning models and data. Our goal is to provide an easy-to-use library comprising  of many checks related to various types of issues, such as 
 model predictive performance, data integrity, data distribution mismatches, and more. The package is distributed under the GNU Affero General Public License (AGPL) and relies on core libraries
from the scientific Python ecosystem: scikit-learn, PyTorch, NumPy, pandas, and SciPy. Source code, documentation, examples, and an extensive user guide can be found at
\url{https://github.com/deepchecks/deepchecks} and \url{https://docs.deepchecks.com/}.

\end{abstract}


\section{Introduction}
Machine learning models are becoming increasingly popular in a variety of fields, including healthcare, finance, biology, and others. Complex models can now be easily trained using modern software packages and yield high predictive performance on test sets. Nevertheless, models are often challenged when deployed outside the lab. 

As indicated in previous works (e.g. \citep{xie2011testing}) detecting faults in machine learning models can be difficult. This is especially true when models are used in sensitive decision-making processes, where mistakes can have serious consequences. 
Despite the sensitivity of the matter, many ML models go into production unmonitored and without proper testing, and thus they are prone to significant risks.

Running ML models in the real world poses a variety of challenges, which may cause degradation in its predictive performance, e.g.:

\begin{itemize}

\item Data integrity issues: Data pipelines are often complex, and the format of the data may change over time. Fields may be renamed, categories may be added or split, and more. Such changes can have a major impact on your model's performance.

\item Data drift and concept drift: Data in the real world is constantly changing. This in turn may affect the distribution of the data that is being fed to the model, or that of the desired target prediction. Thus, the data the model was trained on becomes less and less relevant over time. The format of the data is still valid, but the model will become unstable and its predictive performance will deteriorate over time (in particular in the case of out-of-distribution \citep{geirhos2020shortcu}).
\end{itemize}

\section{The Deepchecks Library: An Overview}
Deepchecks library introduces a framework of data, models, checks, suites, and conditions that enable customizable and extensible testing for ML. It suggests a unified framework with a corresponding API for testing models and data and enables the user to concatenate several checks to be executed later as a single test command.

\begin{figure}[htbp]
\centering
%\includesvg[inkscapelatex=false,width=460pt]{pipeline_when_to_validate.svg}
 \includegraphics[width=460pt]{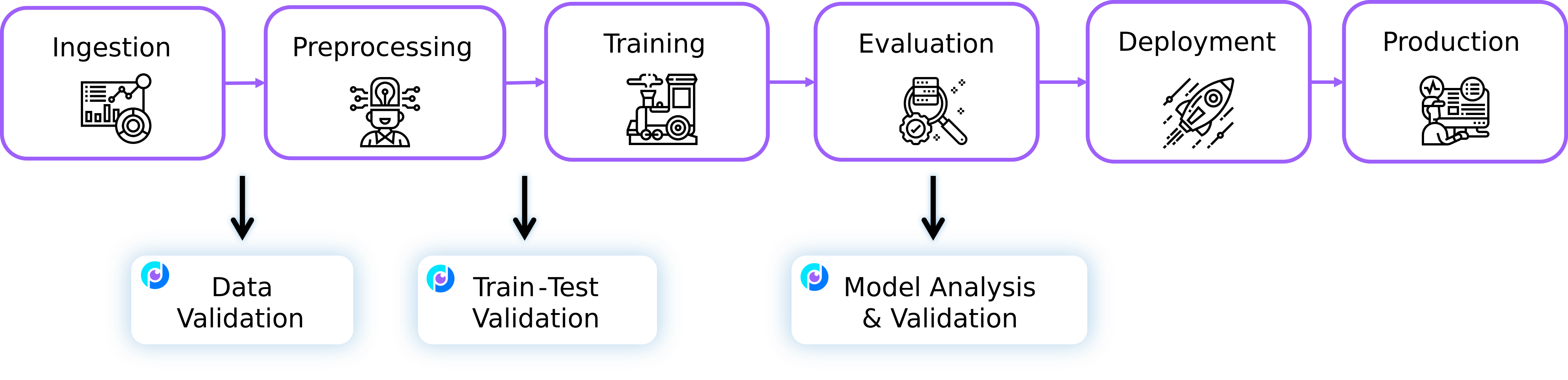}
\caption{When Deepchecks library can be used}
\label{fig1}
\end{figure}

In addition, 
Deepchecks library comes with many built-in checks and suites that can help validate various points throughout the machine learning development process (see Figure \ref{fig1}). 

Of course, every process has its unique steps and challenges, and therefore all checks and suites can be easily customized. In particular, the user can add new checks and their results to support the validation of various phases in the pipeline. Alongside that, we have identified that there are several recurring scenarios, that each has its own needs and characteristics. In particular, deepchecks includes pre-defined suites for the following scenarios:

\begin{itemize}
\item New Data: When a user starts working on a new task, Deepchecks helps to
validate data’s integrity (For example, detecting duplicate samples, problems with string or categorical features, significant outliers, inconsistent labels, etc.)

\item After Splitting the Data (Train-Test Validation) - When splitting the data (e.g. to train, validation or test), and just before training the model, Deepchecks ensures that the splits are indeed representative. For example, it verifies that
the classes are balanced similarly, that there is no significant drift in the distributions between the features or labels in each of the datasets, that there is no potential data leakage that may contaminate the model, etc. 

\item After Training a Model (Analysis and Validation) - Once a trained model is available Deepchecks examines several performance metrics, compares them to various benchmarks, and provides a clear picture about the model’s performance.
Furthermore, Deepchecks attempts to find sub-spaces where the model underperforms and provide insights that may be used to improve its performance.

\end{itemize}

While the above-mentioned scenarios are predefined, a common use case is also to run specific checks on an "on-demand" basis. This is particularly useful when the user is looking into a problem, such as over-fitting.

\section{Deepchecks' Building Blocks}
\begin{figure}[htbp]
\centering
\includegraphics[width = 200pt]{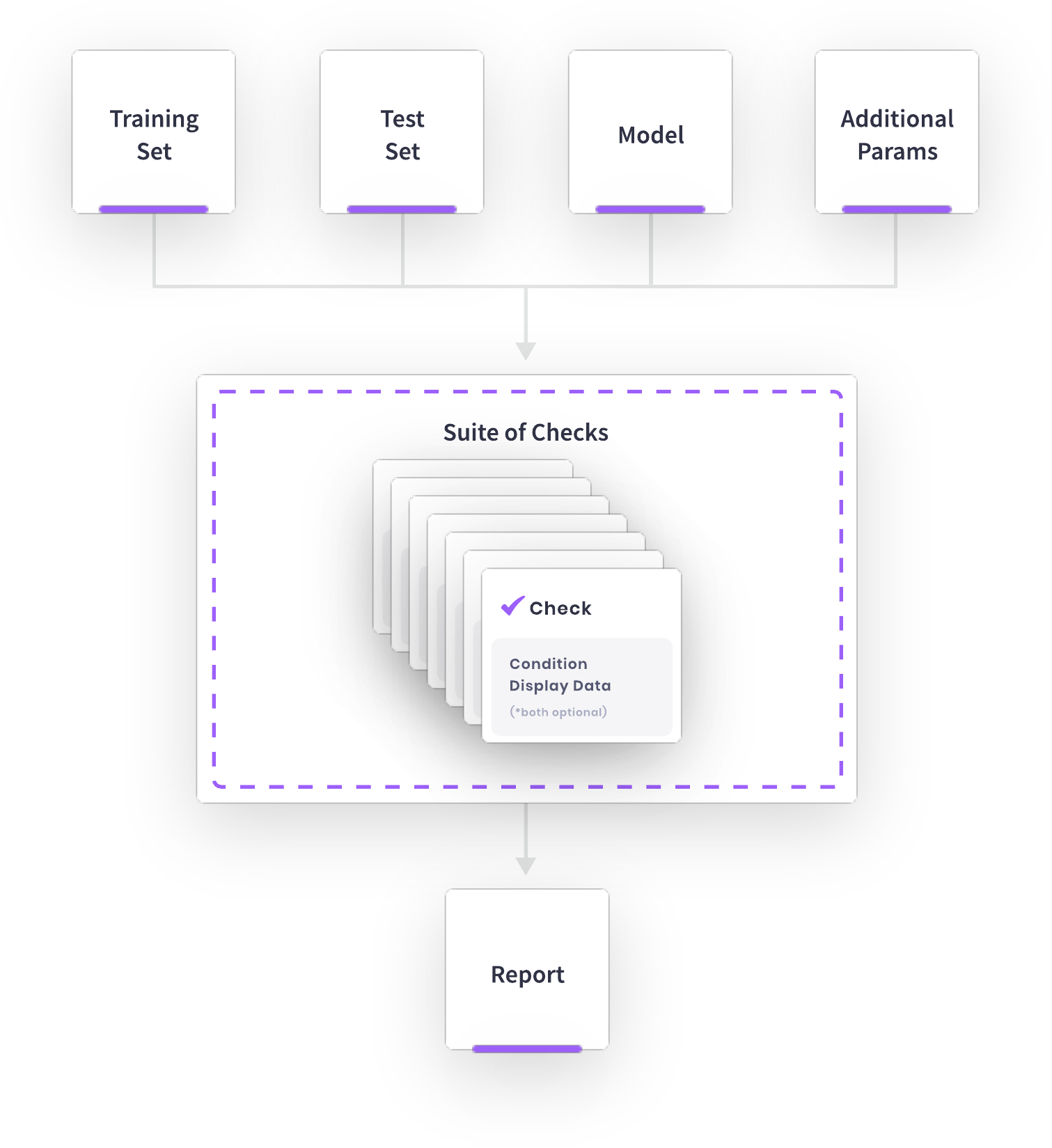}
\caption{The building blocks of Deepchecks library}
\label{fig2}
\end{figure}

Figure \ref{fig2} presents a typical flow of the Deepchecks library. Depending on the checks that the user wishes to execute, some of the following objects should be provided as input: Raw data (before pre-processing), with optional labels, 
the training data (after pre-processing) with the target attribute, test data (which the model is not exposed to), with optional labels, and the model to be checked.

For tabular data, the Deepchecks library requires that the model have a \texttt{predict} method for regression tasks and in addition \texttt{predict\_proba} method for classification tasks, both of which should be implemented using the \texttt{scikit-learn} API conventions \citep{pedregosa2011scikit}. Some checks may attempt using additional model methods if those exist. For example, it uses the built-in feature importance property if it exists, and if it does not, it calculates the feature importance using permutation importance procedure \citep{breiman2001random}. 
Note that built-in scikit-learn classifiers and regressors, along with many additional popular models types (e.g. XGBoost \citep{chen2016xgboost}, LightGBM \citep{ke2017lightgbm}, CatBoost\citep{prokhorenkova2018catboost} etc.) implement these methods and are thus supported.

The three main building blocks of the Deepchecks library are checks, conditions, and suites.

\subsection{Checks}
A check aims to inspect a specific aspect of the data or model. Checks can cover all kinds of common issues, such as data leakage, concept drift, etc. To date, Deepchecks library contains more than 39 checks and supports checks for classification and regression models that are trained on tabular data. The library includes five main categories of checks (called modules):

\begin{enumerate}
\item \texttt{distribution} - this module contains various checks for estimating if the training and test set have different distributions. This module includes checks such as: comparing the model’s trust score \citep{jiang2018trust} of the train and test sets;
calculating the drift between the train set and test set per each input feature (in particular Earth Movers Distance for numerical variables and Population Stability Index for nominal features);
calculating the target attribute drift between train set and test set.

\item \texttt{integrity} - this module contains all data integrity checks, such as: checking for duplicate samples in the dataset; detecting a small amount of a rare data type within a column, such as few string samples in a mostly numeric column; checking if there are columns which have only a single unique value in all rows, etc.

\item \texttt{methodology} - this module contains checks for methodological flaws in the model building process, such as: Checking for overfitting caused by using too many iterations in a gradient boosted model; Detect features that are nearly unused by the model; Calculating the Predictive Power Score\footnote{\url{https://github.com/8080labs/ppscore}} of all features, in order to detect features whose ability to predict the target is due to leakage; Detecting samples in the test data that appear also in training data.

\item \texttt{evaluation} - Module that contains checks of model performance metrics, such as: Comparing the predictive performance of a given model to that of a relatively simple model (e.g. a single tree model) which is used as a baseline for performance comparison.; Calculating the calibration curve with brier score for each class; Checking the distribution of errors; Finding features that best split the data into segments of high and low model error.

\item \texttt{overview} - Module that provides meta-information regarding the model and the dataset, such as:
the role and logical type of each column; the model's hyper-parameters, etc.
\end{enumerate}

\begin{figure}[htbp]
\centering
\includegraphics[width=300pt]{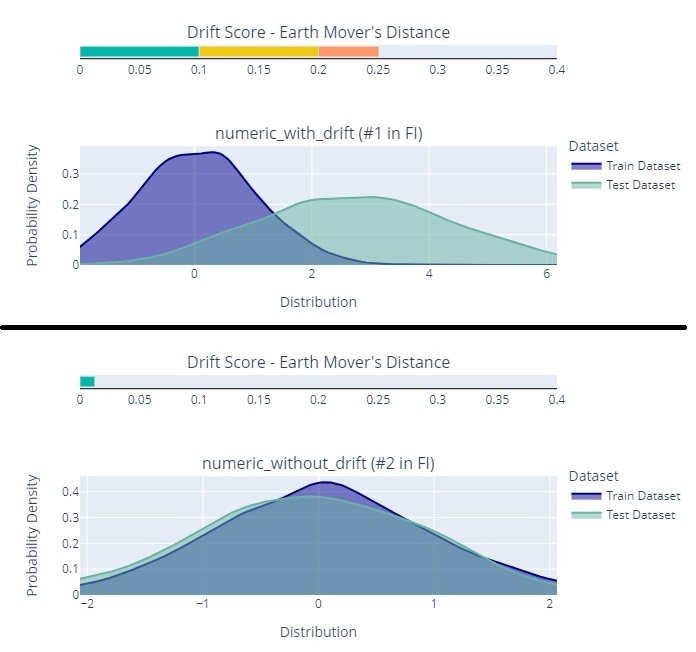}
\caption{An illustration of drift check output. The upper graph shows a feature with relatively high drift score and the lower graph shows a feature with relatively low drift score }
\label{fig3}
\end{figure}

Each check can have two types of results:
\begin{enumerate}
\item A visual result meant for display (e.g. a figure or a table). Figure \ref{fig3} shows one of the outputs that measure the distribution drift between the train set and the test set for two input attributes.

\item A return value that can be used for validating the expected check results. These values can be further used as predicates in conditions (validations are typically done by adding a "condition" to the check, as explained below).
\end{enumerate}

\subsection{Condition}
A condition is a function that can be added to a Check 
for validating if the Check's return value complies with a predefined threshold or logic. A condition returns a status of either pass, fail, or warning result, as well as a statement that describes the status (e.g. ''found 7\% duplicate samples''). The last two results indicate that a flaw may exist and further investigations are required.

For example, a condition attached to the check of \texttt{DataDuplicates} may return the status fail if there are more than 5\% duplicate samples in the training set. This may be valid if the duplicates are there on purpose (e.g., as a result of intentional oversampling, or because the dataset's nature has identical-looking samples), but if this is a hidden issue that is not expected to occur, it may be an indicator for a problem in the data pipeline that needs to be addressed \citep{barz2020we}.

\subsection{Suites}
A suite is an ordered collection of checks, that can have conditions added to them. Once a suite is executed, a summary report is generated which consists of high level results and detailed results. The suite mechanism enables efficiently running a large group of checks with a single call and displaying a concluding report for all of the Checks that ran.
The library comes with a list of common predefined suites for tabular data. The user can build a customized validation scenario, adapted per models, domains, and timing in the pipeline. These suits can be shared with the community and re-used, serving as a framework for methodological testing.

\section{Conclusion and Future Work}
Here, we presented Deepchecks, a library of validating machine learning models and their corresponding datasets. 
The library currently supports classification and regression models for tabular data, and at the time of writing is in beta version for computer vision models trained on images \footnote{\url{https://docs.deepchecks.com/en/latest/examples/vision/checks/}}. We are continuously adding new checks, improving usability, documents, and tutorials.
Finally, we welcome contributors to help us at \url{https://github.com/deepchecks/deepchecks}.

% Acknowledgements should go at the end, before appendices and references

\acks{
We would like to thank everyone on this list for contributing to this project: \url{https://github.com/deepchecks/deepchecks/graphs/contributors}. We would like to thank
the users of Deepchecks library 
for the continuous valid feedback over the last year.}


\newpage

\vskip 0.2in
\bibliography{sample}

\end{document}